\documentclass[10pt,twocolumn,letterpaper]{article}
\pdfoutput=1 
\usepackage{iccv}
\usepackage{times}
\usepackage{epsfig}
\usepackage{graphicx}
\usepackage{amsmath}
\usepackage{amssymb}
\usepackage{enumitem}
\setlist[itemize]{leftmargin=*}
\setlist[enumerate]{leftmargin=*}
\usepackage{hyperref}

\newcommand\nnfootnote[1]{%
  \begin{NoHyper}
  \renewcommand\thefootnote{}\footnote{#1}%
  \addtocounter{footnote}{-1}%
  \end{NoHyper}
}

\iccvfinalcopy 

\def\iccvPaperID{8736} 

\ificcvfinal\fi

\begin{document}

\title{Lightweight Multi-person Total Motion Capture \\ Using Sparse Multi-view Cameras}

\author{Yuxiang Zhang, Zhe Li, Liang An, Mengcheng Li, Tao Yu*, Yebin Liu* \\
Department of Automation and BNRist, Tsinghua University \\
}
\maketitle
\ificcvfinal\thispagestyle{empty}\fi

\begin{abstract}
Multi-person total motion capture is extremely challenging when it comes to handle severe occlusions, different reconstruction granularities from body to face and hands, drastically changing observation scales and fast body movements. 
To overcome these challenges above, we contribute a lightweight total motion capture system for multi-person interactive scenarios using only sparse multi-view cameras.
By contributing a novel hand and face bootstrapping algorithm, our method is capable of efficient localization and accurate association of the hands and faces even on severe occluded occasions.
We leverage both pose regression and keypoints detection methods and further propose a unified two-stage parametric fitting method for achieving pixel-aligned accuracy.
Moreover, for extremely self-occluded poses and close interactions, a novel feedback mechanism is proposed to propagate the pixel-aligned reconstructions into the next frame for more accurate association.
Overall, we propose the first light-weight total capture system and achieves fast, robust and accurate multi-person total motion capture performance. 
The results and experiments show that our method achieves more accurate results than existing methods under sparse-view setups.
\end{abstract} 

\nnfootnote{* Corresponding Author}

\vspace{-2mm}
\section{Introduction}

Marker-less motion capture, due to its great potentials for behaviour understanding, sports analysis, human animation, video editing and virtual reality, has been a popular research topic in computer vision and graphics for decades.
Within this research field, total motion capture, pioneered by ~\cite{joo2018total} using an extremely dense-view setup (hundreds of cameras), shows impressive results of simultaneous capture of multi-person total interactive behaviours including facial expressions, body and hand poses, and has aroused widespread interest in computer vision community.
However, this work \cite{joo2018total} suffers from expensive and sophisticated hardware setup and low run-time efficiency.

\begin{figure}
    \centering
    \includegraphics[width=\linewidth]{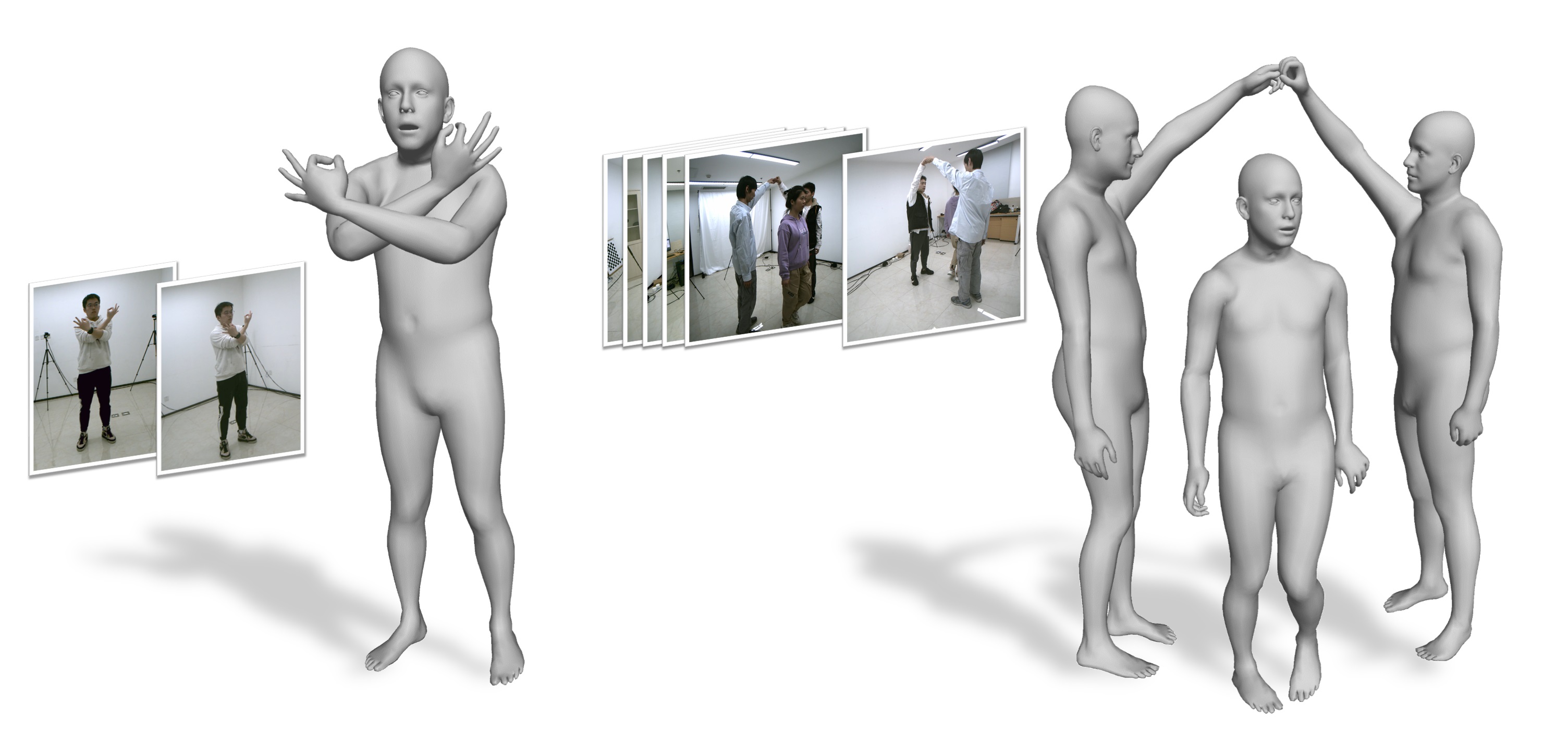}
    \caption{Our lightweight total capture system produces expressive human models with sparse multi-view cameras.}
    \label{fig:teaser}
     \vspace{-2mm}
\end{figure}

Recently, to reduce the capture complexity, more and more researches try to perform total motion capture from only a single image or video~\cite{SMPL-X:2019,xiang2018monoculartotal,MonoExpressive2020,Frankmocap, moon2020pose2pose, zhou2021monocular}. 
By either optimizing a parametric models like SMPL-X~\cite{SMPL-X:2019} and Adam~\cite{joo2018total} (~\cite{xiang2018monoculartotal}) or regressing the model parameters directly from the input images~\cite{MonoExpressive2020}, these methods even achieve real-time total motion capture performance for single-person~\cite{Frankmocap, zhou2021monocular}. 
However, it remains difficult for the monocular methods to handle severe occlusions and challenging poses under multi-person interactive scenarios.

To guarantee both lightweight setups and robust performance, we propose the first lightweight total capture system using only sparse multi-view cameras. However, extending the existing monocular total capture methods to sparse-view multi-person total capture is not trivial. 
Although the incorporation of multi-view observations may resolve the depth ambiguities for monocular methods, the severe occlusions caused by complex-poses and multi-person interactions will significantly deteriorate the performance for current total capture methods. Specifically, the main challenges include: 
i) hand/face association across multiple views under drastically changing observation scales and unstable detection results, 
ii) pixel-aligned fitting between the reconstructed 3D model and the input images,
and iii) robust and accurate body association under severe occlusions even for close interactions. 
To resolve all the challenges above, we propose, as far as we know, the first method to achieve fast, robust and accurate multi-person total motion capture using only light-weight sparse-view cameras. 

First of all, compared with the relatively fixed body part scales and satisfactory occlusion-free view point in monocular single-person total capture cases \cite{SMPL-X:2019,xiang2018monoculartotal,MonoExpressive2020,Frankmocap,zhou2021monocular}, sparse multi-view setups suffer from hand/face fragments on account of severe occlusions, blurs on hands and even fingers by fast limb movements, and varying hand/face scales across different cameras.
Moreover, it remains challenging to associate hands correctly when different hands are located very closely on an image.
To resolve these challenges, we propose a novel hand and face bootstrapping algorithm to extract accurate body part features effectively from the sparse and multi-scale images for accurate association. Benefiting from the recent progress in multi-person skeleton pose capture \cite{4DAssociation}, the skeleton-level results are utilized to guide the following object-detection network for more robust and accurate detection. Moreover, we introduce the cross-modality consistency and cross-scale consistency to filter unexpected detection results of fragments caused by occlusions or improper view points. 



Secondly, using only the pose-regression methods or the key-point detection methods cannot yet guarantee accurate parametric model fitting.
Firstly, pose-regression methods \cite{MonoExpressive2020, Frankmocap, zhou2021monocular} are able to reconstruct decent hand gestures in self-occlusion cases, 
but these one-shot methods cannot guarantee pixel level alignment with 2D joint positions on the image. 
On the other hand, keypoint-detection methods \cite{SMPL-X:2019, xiang2018monoculartotal} are capable of providing pixel-aligned geometric features for visible joints, but may need heavy post-processing optimizations, which is quite sensitive to the initialization and usually fails due to self-occlusion. 
To fully leverage the advantages of both categories and avoid their drawbacks, we propose a new unified two-stage parametric fitting method, in which we leverage the pose-regression result as the initial value to accelerate the convergence for parametric model fitting based on the detected keypoints, and finally achieves pixel-aligned fitting accuracy without losing the efficiency.

Last but not least, for extremely complex poses and close interactions, even 4D association~\cite{4DAssociation} may fail in the body association step, which is an inherent and natural limitation for sparse multi-view setups. To this end, we propose a feedback mechanism in which the reconstructed pixel-aligned human parametric models in the previous frame are propagated into the current frame for enhancing soft visibility information and finally achieve accurate association result.
Benefiting from this novel feedback mechanism, our method is able to capture accurate human behaviours even under scenarios with severe occlusions and close interactions.

Our contributions can be concluded as:
\begin{itemize}
\setlength{\itemsep}{0pt}
\setlength{\parsep}{0pt}
\setlength{\parskip}{0pt}
    \item A new hand and face bootstrapping method that involves the body-level skeleton guidance for more accurate body part localization and self-validated consistency scores to filter out the noise of fragmented detection results by unexpected view points or occlusion observations (Sec.~\ref{sec:bootstrapping}).
    \item A new unified two-stage parametric fitting method that fully utilizes both pose-regression and keypoint-detection methods to produce accurate pixel-aligned 3D human models with expressive motion (Sec.~\ref{sec:fitting}). 
    \item A new feedback mechanism that propagates the accurate reconstruction into the next frame to further improve the association accuracy especially on the severe occluded occasions (Sec.~\ref{sec:feedback}). 
\end{itemize}
\section{Related Work}

\subsection{Total Motion Capture}
Total motion capture methods, which aim at marker-less multi-scale human behaviour capture (including body motion, facial expressions and hand gestures), have shown great potentials in human 4D reconstruction and high-fidelity neural rendering \cite{peng2021neural, sun2021neural, li2021posefusion, zheng2021deepmulticap}.
As the pioneering method of total motion capture, \cite{joo2018total} achieved promising human behaviours capture results under the setup of hundreds of cameras, however, this method relies on the expensive and sophisticated hardware and is therefore hard for applications. On the other end of the spectrum, to achieve lightweight and convenient capture, many works \cite{SMPL-X:2019,xiang2018monoculartotal,MonoExpressive2020,Frankmocap, moon2020pose2pose, zhou2021monocular} focused on total capture from a monocular setup. Monocular total capture \cite{xiang2018monoculartotal} and SMPLify-X \cite{SMPL-X:2019} optimized parametric human models (SMPL-X \cite{SMPL-X:2019} and Adam \cite{joo2018total}) to fit with the 2D detected keypoints. Choutas \textit{et al.} \cite{MonoExpressive2020} directly regressed the parameters of SMPL-X \cite{SMPL-X:2019} from a single RGB image and refined the captured results of head and hands subsequently. Pose2Pose \cite{moon2020pose2pose} combined global and local image features for more accurate prediction. FrankMocap \cite{Frankmocap} regressed parameters of hand and body poses separately and finally integrated two parts into a unified whole body output. Zhou \textit{et al.} \cite{zhou2021monocular} exploited the motion relationship between body and hands to design the network and achieved real-time monocular capture. Overall, although current monocular methods could achieve plausible human total capture performance, they still suffer from depth ambiguity and occlusions. 


\subsection{Skeleton-based Pose Reconstruction}
Single-view 2D and 3D pose estimation methods \cite{wei2016convolutional, pishchulin2016deepcut,he2017mask, fang2017rmpe, cao2018openpose, Chen2018CPN,  li2018crowdpose, zanfir2018deep, alp2018densepose, mehta2018single, kanazawa2018hmr, nie2019single, mehta2020xnect, pymaf2021} have been widely explored in recent years, however, they suffer from severe occlusions and ambiguity and cannot produce high-confidence results. To alleviate the occlusion and produce more accurate reconstruction, many works aimed to reconstruct human poses from multi-view input. On the first branch of this direction, some approaches \cite{gall2009motion, taylor2010dynamical, liu2011markerless, yao2011learning, liu2013markerless, li2018PG, kwon2020recursive, ohashi2020synergetic} preformed temporal skeleton-based tracking for each frame, but these methods suffer from imperfect initialization and accumulated errors. On another branch, cross-view matching methods associated correspondences (e.g., human instances and keypoints) from different viewpoints and finally reconstructed 3D pose for each performer. Some works utilized 3DPS models to solve 3D joint positions implicitly by skeletal constraints \cite{belagiannis20143d,  belagiannis20163dps} or body part detection \cite{ershadi2018multiple}. Joo \textit{et al.} \cite{joo2017panoptic} utilized 2D detection from dense multiple views to vote for possible 3D joint positions. Dong \textit{et al.} \cite{dong2019fast} proposed a multi-way matching algorithm to guarantee cycle consistency across all the views. Zhang \textit{et al.} \cite{4DAssociation} jointly formulated the temporal tracking and cross-view matching as a 4D association graph and achieved real-time performance. Tu \textit{et al.} \cite{tu2020voxelpose} proposed to directly operate in the 3D space while avoiding incorrect decisions in each viewpoint. Lin \textit{et al.} \cite{lin2021multi} presented a plane-sweep-based approach to perform multi-view multi-person 3D pose estimation without the explicit cross-view matching. Even though these methods are able to capture 3D human poses using skeletons, they cannot reconstruct full body behaviours, i.e., facial expressions, hand motions, and body surfaces. 

\subsection{3D Hand Reconstruction}
3D hand reconstruction is an essential sub-problem in total capture. Many works \cite{simon2017hand, cai2018weakly,spurr2018cross,yang2019aligning,zimmermann2017learning,iqbal2018hand,mueller2018ganerated} that focused on 3D hand pose estimation from a single RGB image have been proposed. Recently, more and more works aimed at recovery of a 3D hand mesh \cite{ge20193d,kulon2020weakly,chen2021camera} or directly regressing the pose and shape parameters of a parametric hand model (MANO \cite{romero2017embodied}) \cite{baek2019pushing,boukhayma20193d,zhang2019end,zimmermann2019freihand,chen2021model}. However, these methods only focused on single hand reconstruction while ignoring the interaction between hands. Moon \textit{et al.} \cite{moon2020interhand2} proposed InterHand2.6M, a large-scale two hand interaction dataset. Several researchers have explored the problem of pose estimation under two hand interacting situation \cite{mueller2019real,wang2020rgb2hands,Lin_2021_WACV,wang_2021_ICCV}, 
but the problem of hand pose estimation under multi-person interacting scenario with more hands involved is still unsolved.

\section{Overview}
\label{sec:formulation}
\begin{figure*}[ht!]
    \centering
    \includegraphics[width=\linewidth]{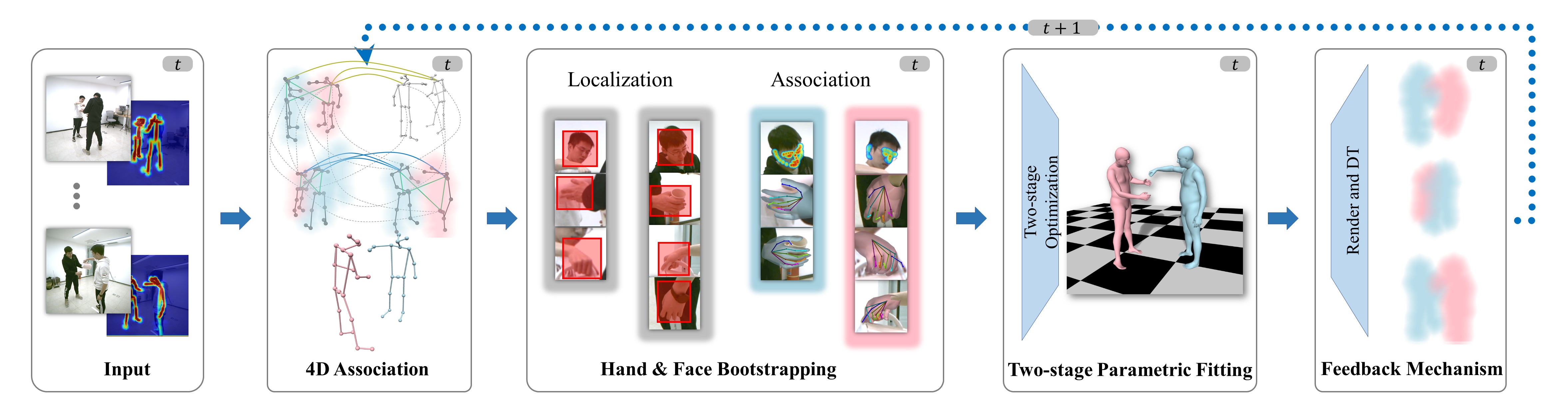}
    \vspace{-3mm}    
    \caption{Method overview. Initially, we take multi-view RGB sequences and body estimation results as our inputs. Skeletons of each individuals are constructed by 4D association(Sect.~\ref{sec:overview:4D_association}). After that, we utilize our limb bootstrapping framework to localize(Sect.~\ref{sec:sec:hand_localization}) and associate(Sect.~\ref{sec:sec:hand_association}) body part. After that, we optimize parametric SMPL-X models from all these outputs(Sect.~\ref{sec:fitting}). Finally, our feedback mechanism(Sect.~\ref{sec:feedback}) is introduced to boost the body association performance in next frame with the reconstructed human model.}
    \label{fig:overview_pipeline}
    \vspace{-3mm}
\end{figure*}

\subsection{Main Pipeline}
\label{sec:main pipeline}
As shown in Fig.~\ref{fig:overview_pipeline}, given multiple synchronous and calibrated RGB videos as input, our pipeline works in a frame-by-frame manner, and outputs a series of parametric human models naturally combining body posture, hand gesture and facial expressions by the following steps:
\begin{enumerate}
	\setlength{\itemsep}{0pt}
	\setlength{\parsep}{0pt}
	\setlength{\parskip}{0pt}
\item \textbf{4D Body Association} (Sec.~\ref{sec:overview:4D_association}): 
Given multi-view input, we associate the 2D keypoints and triangulate 3D body skeletons using 4D association \cite{4DAssociation}.
\item \textbf{Hand and Face Bootstrapping} (Sec.~\ref{sec:bootstrapping}): With the body skeletons, we perform hand and face bootstrapping to extract their 2D bounding boxes efficiently and also associate them to different subjects among different views. 
\item \textbf{Two-stage Parametric Fitting} (Sec.~\ref{sec:fitting}): Then we fit parametric human model SMPL-X~\cite{SMPL-X:2019} to these posture, gesture and expression features in a two-stage manner to achieve efficient and accurate pixel-level alignment.
\item \textbf{Feedback Mechanism} (Sec.~\ref{sec:feedback}): Finally, the tracked human models are propagated into the 4D association step of the next frame to further improve the association accuracy especially on severe occluded occasions.
\end{enumerate}

\subsection{4D Body Association}
\label{sec:overview:4D_association}
As a building block of our method, the 4D association \cite{4DAssociation} contributes a real-time multi-person skeleton tracking framework with sparse multi-view video inputs. 
By taking the tracked 3D joints from the previous frame and the detected 2D key-points in current frames as graph nodes $\mathcal{D}_j$, 4D association algorithm introduces a series of connecting edges: single-view parsing edges $\mathcal{E}_P$, cross-view matching edges $\mathcal{E}_V$ and temporal tracking edges $\mathcal{E}_T$, and finally formulate a unified association graph $\mathcal{G}_{4D}$ for optimizing the multi-view body association problem effectively. 


\section{Hand and Face Bootstrapping}
\label{sec:bootstrapping}

We introduce a hand and face bootstrapping method to (i) extract local body part regions of interest (RoI) and detection from full-body inputs and (ii) eliminate incorrectly associated matches using the proposed non-maximum suppression (NMS) method. Body-level semantic features, hand pose regressions and keypoint detections are integrated into our pipeline. 
Note that the proposed bootstrapping methods for hand and face are quite similar, but the interactive hand behaviour is much more frequent under practical multi-person scenarios. So in this section, we mainly introduce the hand bootstrapping method which is more representative, and the method for face is similar.


Specifically, given sparse multi-view image inputs at frame $t$, we firstly leverage the 4D association algorithm (Sec.~\ref{sec:overview:4D_association}) to get the associated 2D body keypoints in each view and the triangulated 3D body skeletons.
Secondly, we indicate preliminary screened RoIs $\{{RoI}^c_{\alpha}\}$ through body skeleton semantic information, then a lightweight object-detection network is utilized for further localizing tight and reliable RoIs $\{{RoI}^c_{\beta}\}$ in these initiatory screened areas to boost the key point detection and parametric regression performance of hands. However, there may exist several ${RoI}^c_{\beta}$ corresponding to different hands in a single ${RoI}^c_{\alpha}$ due to close interactions and bad view directions as shown in Fig.~\ref{fig:hand_association} (b). This will lead to severe ambiguity in the later hand association step. To eliminate these ambiguous RoIs, we propose a double-check non-maximum suppression (NMS) method to guarantee both \textbf{cross-modality} (between key point detection and parametric regression of hands) and \textbf{cross-scale} (between body reconstruction and hand reconstruction) consistency. 
Next, we will introduce the 2D hand localization and association in detail.


\subsection{2D Hand Localization}
\label{sec:sec:hand_localization}
We conduct 2D hand localization in a coarse-to-fine manner: first generating initial bounding box for each hand according to the reconstructed 3D body skeleton and semantic information, and then refine the initial bounding boxes using the iterative hand detector~\cite{Wang:2019:SRH}. Note that this strategy helps us filter out the inconsequential areas efficiently at the coarse level, thus reducing unnecessary computation and accelerating the hand localization process. 

For generating the initial bounding box for a hand, we leverage the reconstructed 3D body skeleton to interpolate the center of hand and construct a 3D bounding sphere with constant radius to handle size variations of hands on 2D images caused by perspective projection. 
We then generate the initial 2D bounding box according to the projected center and radius of the 3D bounding ball in each view. 
Specifically, we estimate $\{{RoI}^c_{k,\alpha}\}$ of person $k$ in each view $c$ under the guidance of the reconstructed body skeletons: 
\begin{equation}
\begin{split}
    o^c_p &= P_{c}(O_p), \quad r^c_p = \frac{f_c \cdot R} {d_{c}(O_p)}, \\
    \{{RoI}^c_{k,\alpha}\} &= \{Rect(o^c_p, r^c_p)| z^c_p = 1 ,p=1,2,...,P\} ,
\end{split}
\end{equation}
where $O_p$ and $R$ is sphere center and radius. $O_p$ could be simply extrapolated from the 3D position of wrist and elbow, and $R$ is a constant parameter defined in terms of realistic physical scale. $o^c_p$ and $r^c_p$ are the projected circle center and radius in view $c$, respectively. $f_c$ is the focal length of camera $c$, $P_{c}(\cdot)$ is perspective projection function, and $d_{c}(\cdot)$ is the distance to camera's image plane. An indicated variable $z^c_p$ is introduced for whether the wrist joint of person $p$ has been assigned a 2D keypoint detection in view $c$. 



\begin{figure}
  \centering
  \includegraphics[width=\linewidth]{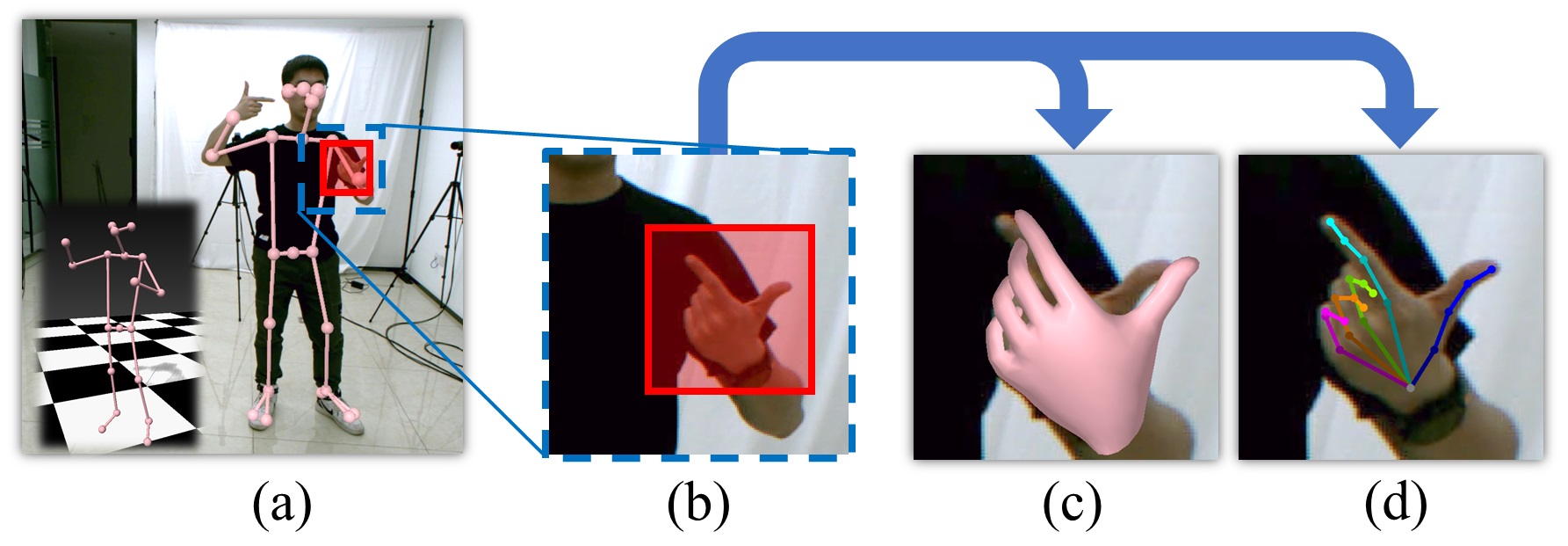}
  \caption{Illustration of hand localization and detection. (a) Reconstructed body skeletons using \cite{4DAssociation} in advance which will guide us to focus on key areas $\{{RoI}^c_{k,\alpha}\}$ (blue dotted line rectangle). 
  (b) After that, a light-weight network is utilized for regressing more precise and tight bounding boxes $\{{RoI}^c_{k,\beta}\}$(red solid line rectangle).
  Then we clipped $\{{RoI}^c_{k,\beta}\}$ from the full-body image and then feed them to both the pose-regression network and the keypoint-detection network. (c) The regressed hand gesture which is decent but not pixel-aligned accurate enough. (d) 2D Keypoints are accurate but suffer from depth ambiguity. }
  \label{fig:hand_localization}
  \vspace{-3mm}
\end{figure}

Since current hand keypoints detection and regression methods still rely on tight and accurate bounding boxes for achieving good performance, we further refine the initial bounding boxes using the iterative hand detector~\cite{Wang:2019:SRH}. 
As shown in Fig.~\ref{fig:hand_localization}, we utilize that one-pass hand detection network to further extract more precise RoIs $\{{RoI}^c_{k,\beta}\}$. We demonstrate that our two-step localization method outperforms the light-weight detector (e.g., 100DOH \cite{shan2020understanding} used in FrankMocap \cite{Frankmocap}) on both the speed and accuracy (Fig.~\ref{fig:single_qualitative}).

\subsection{Hand Association}
\label{sec:sec:hand_association}

Since 2D body joints have been associated and 3D body skeletons of different subjects have been triangulated in previous steps, in this section, we mainly focusing on how to assign correct bounding boxes of hands to the 3D wrist joints in each view. 

\begin{figure}[t]
  \centering
  \includegraphics[width=\linewidth]{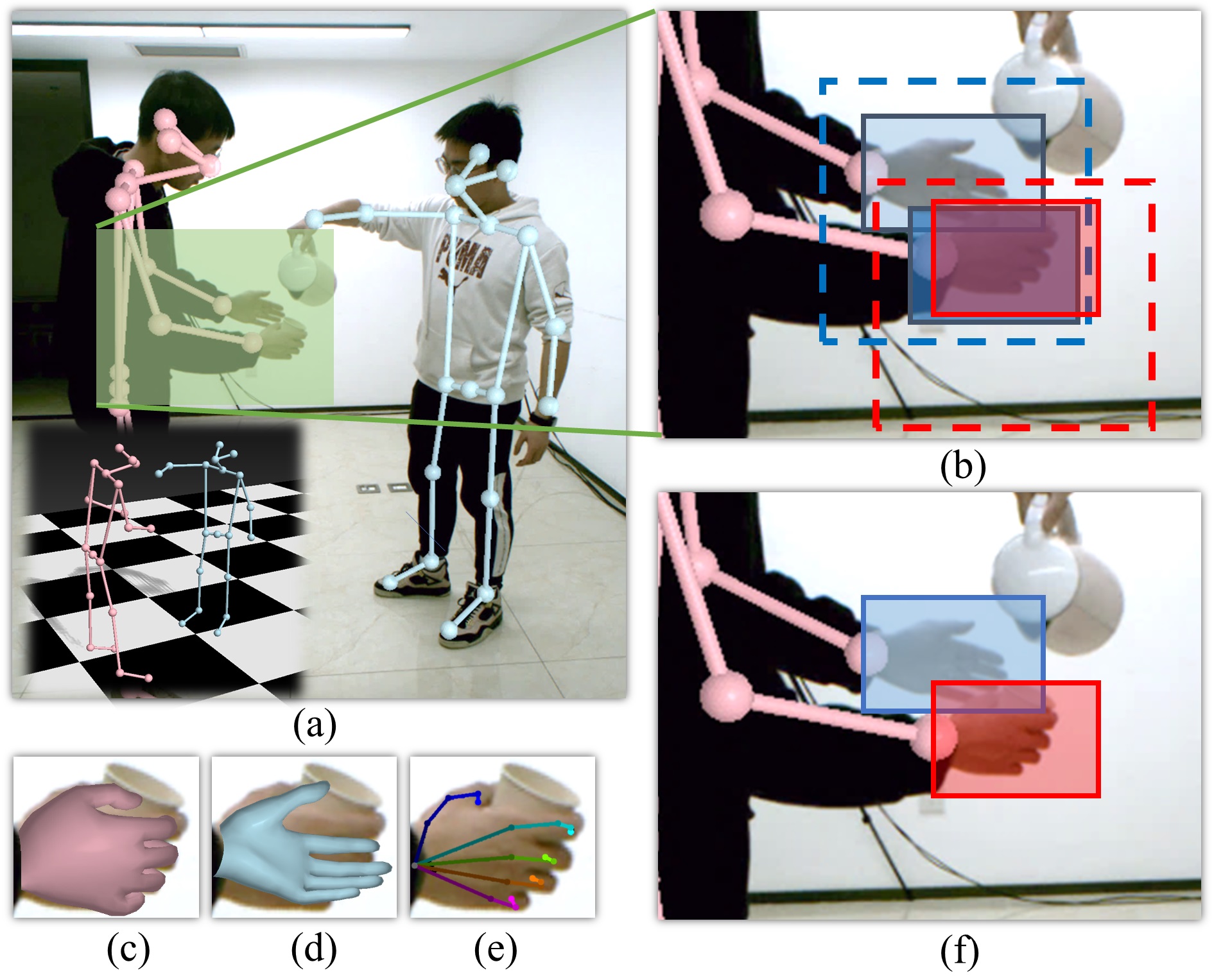}
  \caption{Illustration of hand association algorithm. (a) Body skeletons obtained by \cite{4DAssociation}. (b) Association ambiguity may happen when ${RoI}^c_{k_1,\alpha} \cap {RoI}^c_{k_2,\alpha} \neq \varnothing, k_1 \neq k_2$. Blue and red dotted line rectangles are ${RoI}^c_{k_1,\alpha}$ and ${RoI}^c_{k_2,\alpha}$, respectively. Then 3 tight bounding boxes ${RoI}^c_{k_1,\beta_1}$ (light blue), ${RoI}^c_{k_1,\beta_2}$ (dark blue) and ${RoI}^c_{k_2,\beta}$ (red) are further extracted from the two initial rectangles. We can observe that the right hand lies in the overlapping area of ${RoI}^c_{k_1,\alpha}$ and ${RoI}^c_{k_2,\alpha}$, leading to redundant proposals and confusing partition. (c) and (d) show that pose-regression network is specific for one chirality input. (e) is the result of heatmap-based detector which is invariant with chirality. (f) shows that after our double-check NMS procedure, the correct distributed ${RoI}^c_{k_1,\beta_1}$ are retained and the false one ${RoI}^c_{k_1,\beta_2}$ is discarded.} 
  \label{fig:hand_association}
  \vspace{-3mm}
\end{figure}

We leverage classical non-maximum suppression (NMS) \cite{2017Soft} algorithm but proposed two novel consistency scores to effectively filter out ambiguous RoIs. 
Specifically, cross-modality consistency score ${\zeta}^c_k$ and cross-scale consistency score ${\xi}^c_k$ are proposed to judge which match will be finally retained. 
In practice, hands usually come close or even overlapped in some side views, and interactions among individuals will lead to more ambiguities. Specifically, considering 
the case that ${RoI}^c_{k_1,\alpha} \cap {RoI}^c_{k_2,\alpha} \neq \varnothing, k_1 \neq k_2$, as illustrated in Fig.~\ref{fig:hand_association}, ${RoI}^c_{k_1,\beta_2}$ (dark blue) and ${RoI}^c_{k_2,\beta}$ (red) share the same sub-region, which results in association ambiguities. Inspired by traditional NMS algorithms in object-detection pipeline, we come up with a self-validated association algorithm to filter out redundant ${RoI}^c_{\beta}$ and retain the correct match. Firstly, for each view $c$, we locate all redundant RoIs by calculating Intersection of Union (IoU) from total individuals' hands proposals. Secondly, we calculate \textbf{cross-modality consistency} score ${\zeta}^c_k$ and  \textbf{cross-scale consistency} score ${\xi}^c_k$ for each RoI proposal. 

The first metric, cross-modality consistency score ${\zeta}$, is used to penalize the inconsistency between different detection modalities. As shown in Fig.~\ref{fig:hand_association} (c), (d) and (e), the heatmap-based feature is invariant with respect to flipping translation, but pose-regression network needs right chirality assurance for achieving reasonable results. This divergence can help us to distinguish the left or right association ambiguities. Denote $J_h = 21$ as the hand joint number, $S^h_{regr} \in \mathbb{R}^{2\times J_h}$ as 2D hand joint positions from pose-regression network, $S^h_{dect}\in \mathbb{R}^{2\times J_h}$ as the output of keypoint-detection network, and $w$, $h$ as the size of ${RoI}_{\beta}$. Then ${\zeta}$ is formulated as

\begin{equation}
    {\zeta} = \frac{1}{J_h} \sum_{j=1}^{J_h} max(0, 1 - \frac {2\Vert S^h_{regr, j}-S^h_{dect,j} \Vert_2}{\sqrt{w^2 + h^2}}).
      \label{eqn:cross_modality}
\end{equation}

On the other hand, the second metric, cross-scale consistency score ${\xi}$, is formulated to punish unreasonable wrist misalignment between the local hand estimator and global body estimator. Denote $S^b_{wrt}\in \mathbb{R}^2$ as the associated 2D wrist position from full-body detections, $S^h_{dect,j_w}$ as wrist joint position by local hand keypoints detector. Finally, ${\xi}$ is defined as
\begin{equation}
    {\xi} = max(0, 1 - \frac {2\Vert S^b_{wrt}-S^h_{dect,j_w} \Vert_2}{\sqrt{w^2 + h^2}}).
    \label{eqn:cross_scale}
\end{equation}
Finally, we sum up these two scores as confidence measurements to apply the NMS algorithm to reserve the one with the highest score. 
We demonstrate that our double-check NMS method helps to improve the association accuracy in confusing situations. 
\section{Two-stage Parametric Fitting}
\label{sec:fitting}
We observe that previous methods usually utilize either parametric pose regression \cite{Frankmocap} or heatmap-based keypoints \cite{SMPL-X:2019} for total motion capture. However, they do have their own limitations. 
On one hand, although pose-regression networks can produce plausible results even under occlusions, they can not guarantee accurate 2D alignment with the input image. 
On the other hand, heatmap-based networks provide accurate 2D detections for visible joints, but they are still suffer from depth ambiguities and are susceptible to local minima during optimization. 
In this paper, we unify them together in a two-stage parametric fitting scheme, which contains local initialization and total optimization, to boost the total motion capture performance. 

\noindent\textbf{Local Initialization}
To accelerate convergence and prevent optimization deviation, it is essential to initialize the motion of each body part to a reasonable status. Specifically, for hand pose initialization, we pick a decent initial value from the semantic pose-regression gestures according to the hand association score ${\zeta}$ (Eqn.~\ref{eqn:cross_modality}) and ${\xi}$ (Eqn.~\ref{eqn:cross_scale}).
Besides, for body/head pose initialization, we solve the SMPL-X body pose by minimizing the following energy function directly to guarantee more accurate initialization:
\begin{equation}
    E_{body} = \lambda_{b3d} E_{b3d} + \lambda_{pri} E_{pri} + \lambda_{\beta} E_{\beta}
\end{equation}
Here, $E_{b3d}$ is the distance from parametric model's joints to the corresponding reconstructed 3D body skeletons. $E_{pri}$ and $E_{\beta}$ are used to regularize natural pose and shape optimization as in SMPLify-X~\cite{SMPL-X:2019}. 
        
\noindent\textbf{Total Optimization}
In this stage, we leverage the accurate 2D hands keypoints and faces landmarkers to further optimize the initial SMPL-X model for accurate total capture:
\begin{equation}
    \begin{split}
    E_{total} &= E_{data} + E_{reg}, \\
    E_{data} &= \lambda_{b3d} E_{b3d} + \lambda_{h2d} E_{h2d} + \lambda_{f2d} E_{f2d}, \\
    E_{reg} &= \lambda_{pri} E_{pri} 
    +\lambda_{\beta} E_{\beta} +\lambda_{\theta,h} E_{\theta,h} 
    + \lambda_{\varepsilon} E_{\varepsilon},
    \end{split}
\end{equation}
where $E_{h2d}$ and $E_{f2d}$ are 2D data terms to minimize the distances between the 2D projections of the SMPL-X joints and the detected 2D keypoints in all the valid viewpoints. 
$E_{\theta,h}$ and $E_{\varepsilon}$ are L-2 norm to keep the optimized gestures and expressions within reasonable ranges. 
Note that we can additionally leverage the consistency scores ${\zeta}$ and ${\xi}$ (Eqn.~\ref{eqn:cross_modality} and~\ref{eqn:cross_scale}) to balance the detection results in different views, so $E_{h2d} = \Sigma_{c}\frac{{\zeta^c} + {\xi^c}}{2} \cdot e^c_{h2d}$, where $c$ is the view index.

\begin{figure}
  \centering
  \includegraphics[width=\linewidth]{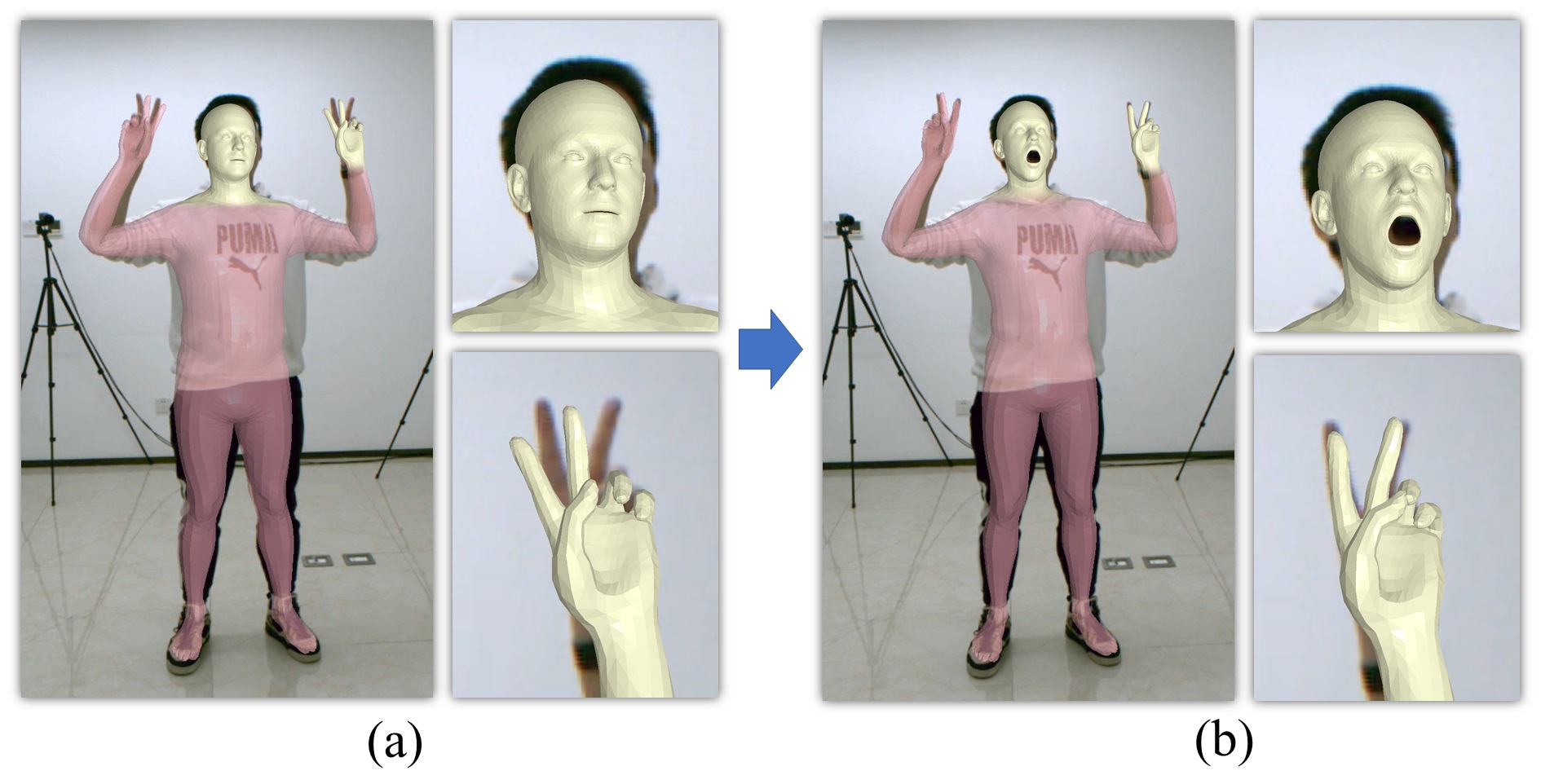}
  \caption{Illustration of two-stage parametric fitting. (a) Stage 1: we solve the body posture as well as arm kinematics and assign the gesture of pose regression with the highest association score ${\zeta}$ and ${\xi}$. (b) Stage 2: we then perform total optimization to achieve accurate total motion capture. } 
  \label{fig:two_stage_optimization}
  \vspace{-3mm}
\end{figure}

\section{Feedback Mechanism}
\label{sec:feedback}

\begin{figure}
  \centering
  \includegraphics[width=\linewidth]{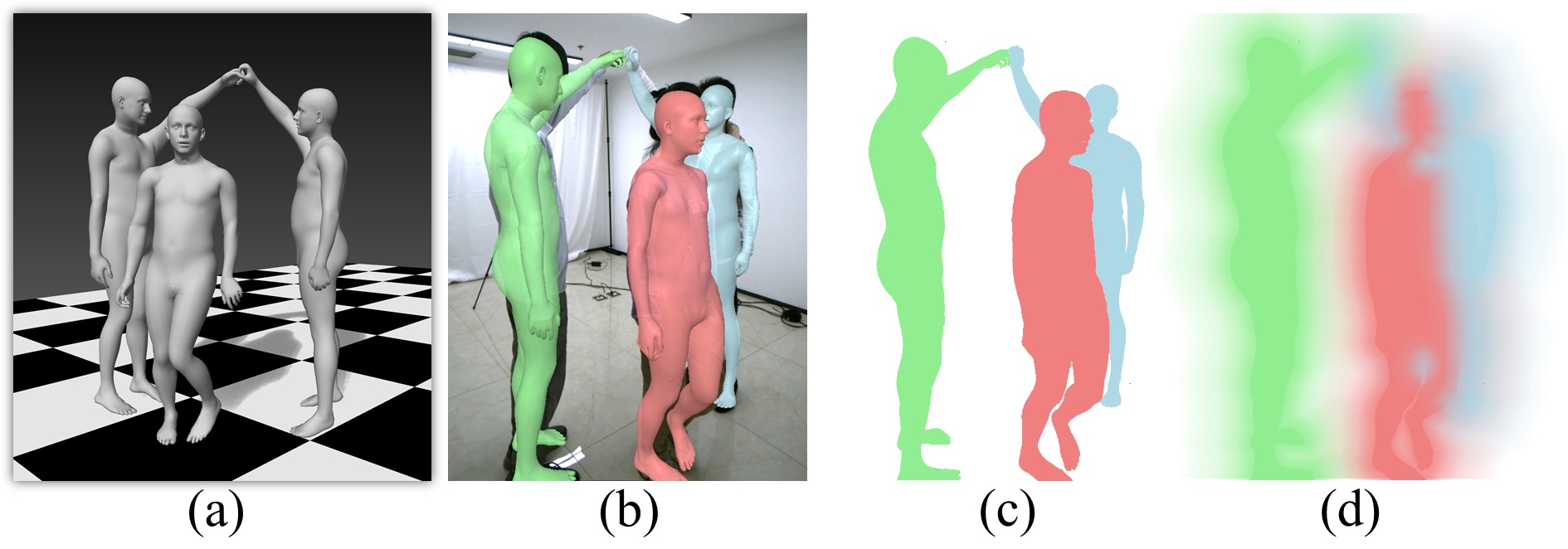}
  \caption{Illustration of our feedback mechanism. (a) and (b) are the aligned parametric models. (c) Segmented results by rendering. (d) is the softened mask generated by distance transformation to enhance the robustness of association during fast motion.} 
  \label{fig:feedback_soft_mask}
  \vspace{-3mm}
\end{figure}

Last but not least, for severe occlusions and close interactions, we put forward a feedback mechanism to boost the tracking performance of the association algorithm in return. On one hand, detailed limb detector contributes to extremity reconstruction with higher precision, which are leveraged to refine the body skeleton results. On the other hand, we re-render the human model to each view for the next frame to extend the tracking edges ${\mathcal{E}}_T$ of $\mathcal{G}_{4D}$ with additional visibility information. As shown in Fig.~\ref{fig:feedback_soft_mask}, we obtain the initial segmentation by rendering the optimized parametric models back to input images. Meanwhile, in order to enhance the robustness with body movements, we implement distance transformation to smooth the boundary of the rendered mask. 

For a given 2D keypoint detection candidate $c$, we use the same denotation $z^k(c)$ in \cite{4DAssociation} to refer the possibility to connect that candidate to person $k$. Benefiting from our feedback module, the tracking edges in 4D association \cite{4DAssociation} (Sec.~\ref{sec:overview:4D_association}) are extended with visibility priors. We define the enhanced tracking edges $\widehat{z}^k(c)$ as:

\begin{equation}
\widehat{z}^k(c) = \frac{{\tau}^k(c)}{ \sum_{i=1}^{K} {\tau}^i(c)} z^k(c),
\end{equation}
where ${\tau}^i(c)\in [0, 1]$ means the continuous occupancy of person $i$. As shown in Fig.~\ref{fig:feedback_soft_mask} (d), ${\tau}^i(c)$ is negative interrelated to the distance to its binary mask, ${\tau}^k(c) = 1$ refers to the fully contained situation. As a result, our feed-back mechanism enhances skeleton tracking performance and reduces jittering in projection coincidence cases.

\begin{figure*}
\centering
  \includegraphics[width=0.9\linewidth]{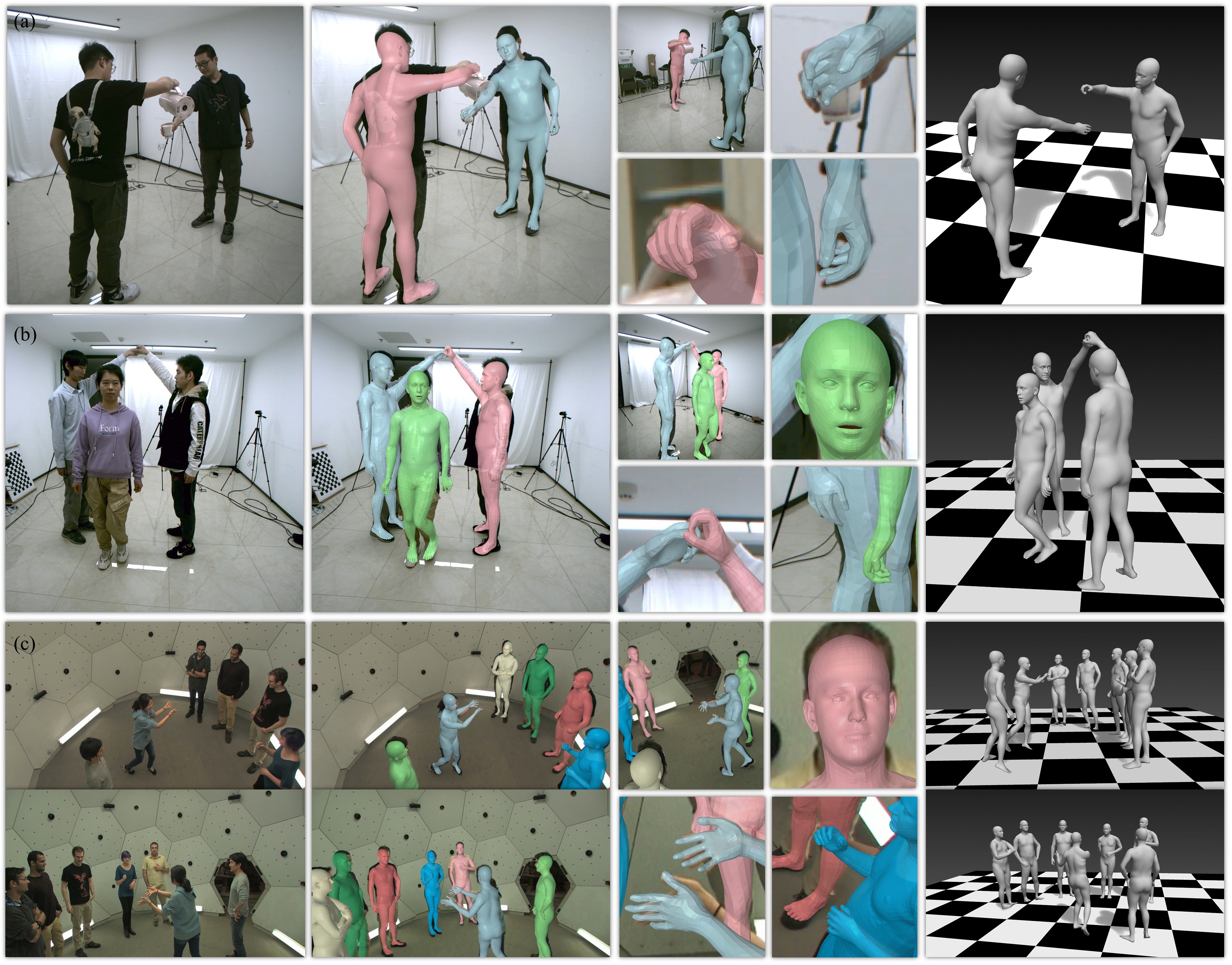}
  \caption{Results by our system. From the left to right are input reference images, parametric model alignment, facial and hand alignment and 3D visualization from a novel view, respectively.
  (a) Results of the hand-object-interaction case from our captured data using 6 views, (b) results of a multi-person-interaction scenario using 6 views, (c) results on CMU dataset ~\cite{joo2018total} using 8 views.
  }
\label{fig:results}
\end{figure*}

\section{Results}
\label{sec:results}
In Fig.~\ref{fig:results}, we demonstrate example results by our system. With the sparse multi-view setup, our method produces expressive human parametric models under multi-person interactive scenarios.

\begin{figure}
  \centering
  \includegraphics[width=0.95\linewidth]{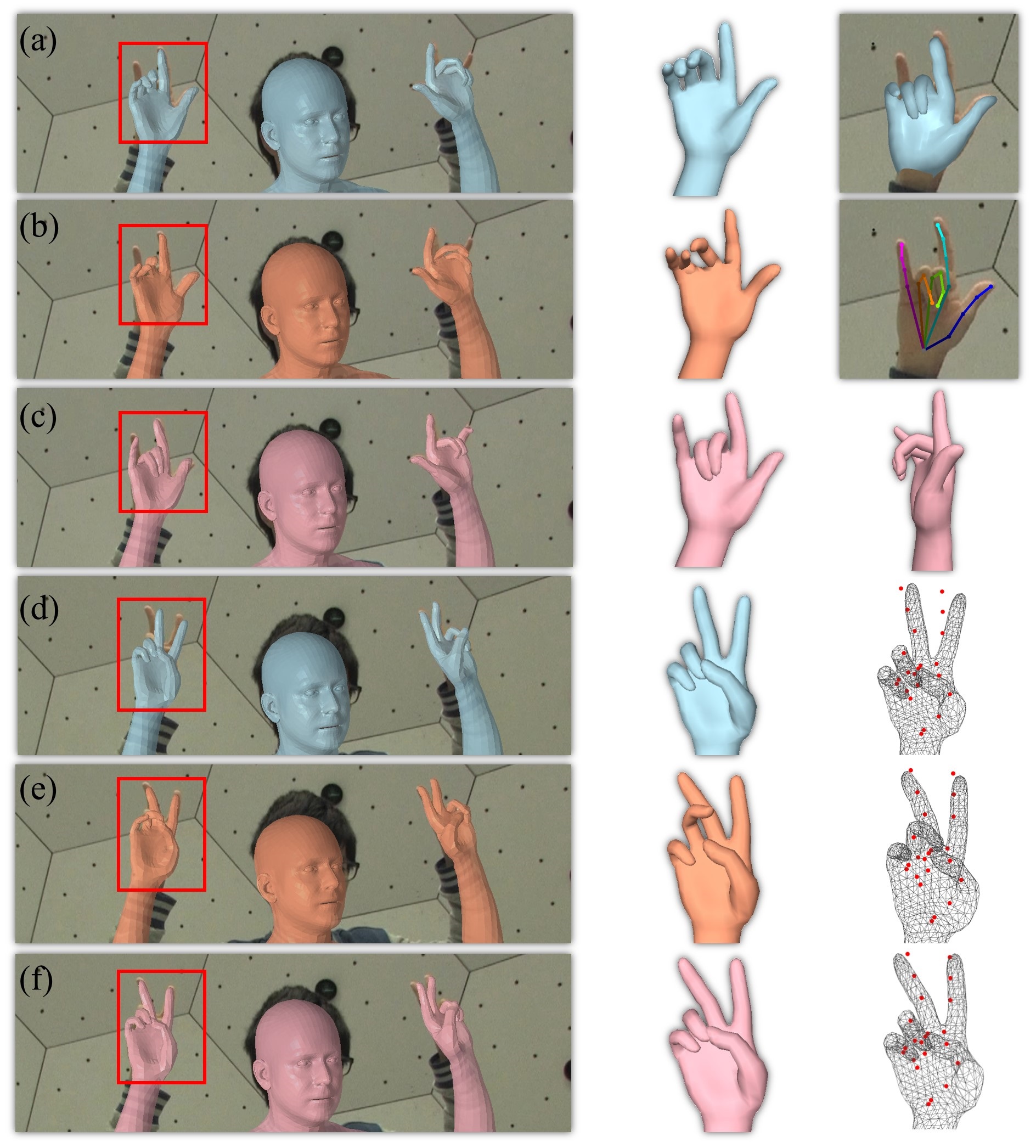}
  \caption{Qualitative evaluation of two-stage parametric fitting. (a) and (d) are the results with parametric-regression-only metric (blue). (b) and (e) are the results with keypoints-detection-only metric (salmon). (c) and (f) are the results with our two-stage fitting strategy combing both metrics (pink). Meanwhile, we visualize pose-regression network outputs and heatmap-based network outputs in right of (a) and (b), respectively. Red dots in right of (d) (e) (f) refer to the ground-truth 3D hand annotations. }
  \label{fig:qual_fitting}
\end{figure}

\subsection{Implementation Details}
\label{sec:system}
Our light-weight total capture system are implemented with 6 synchronized RGB cameras (resolution $2048\times 2048$) on a single PC (i5-6600K CPU, NVIDIA RTX 3090 GPU). We use Openpose \cite{cao2018openpose} as our body pose estimator, SRHandNet \cite{Wang:2019:SRH} as our hand instance detector and keypoint detector. We leverage the hand pose-regression network of Frankmocap \cite{Frankmocap} gesture regressor. FaceAlignment \cite{bulat2017far} are used for face keypoints extraction. Besides, we accelerate all neural network inference by implementing half-precision arithmetic on NVIDIA TensorRT platform. The CNN performance is shown in Table.~\ref{tab:implementation}. Besides, our body association backbone takes nearly 10ms to recover human skeletons, the limb localization and association method is fast enough to be neglected. Our parametric fitting workflow takes 150 ms for stage one and 350 ms for stage two (20 Gauss-Newton iterations and parallel for each person). On the whole, our system run-time depends on the captured individual number and view number. Empirically, our pipeline runs about 1 fps for 2 person with 6 views, and the processing speed slows down to 0.3 fps for 7 persons with 8 views. As for hyper-parameters, sphere radius $R$ in hand localization is set to $0.15m$, and association NMS threshold is 0.5. In two-stage parametric fitting, we set $\lambda_{b3d} =10$, $\lambda_{h2d} =0.0001$, $\lambda_{f2d}=0.0003$,
$\lambda_{pri}=\lambda_{\theta,h}=0.01$, and $\lambda_{\beta}=\lambda_{\varepsilon}=0.01$. 

\begin{table}[ht]
\centering
\begin{tabular}{lllll}
\hline
\quad Network &\quad Input &Batchsize &Speed(FPS) \\ \hline
Openpose~\cite{cao2018openpose} &$368\times368$ &\qquad 6 &\quad 43.1 \\
FaceAlignment~\cite{bulat2017far} &$256\times256$ &\qquad 4 &\quad 109.5 \\ 
SRHandNet~\cite{Wang:2019:SRH} &$256\times256$ &\qquad 8 &\quad 50.0 \\ 
HandHMR~\cite{Frankmocap} &$224\times224$ &\qquad 8 &\quad 202.1 \\ \hline
\end{tabular}
\vspace{2mm}
\caption{Inference speed of the CNN networks used in our system.}
\label{tab:implementation}
\end{table}

\subsection{Comparison}
\label{sec:comparison}
\begin{figure}
  \centering
  \includegraphics[width=0.8\linewidth]{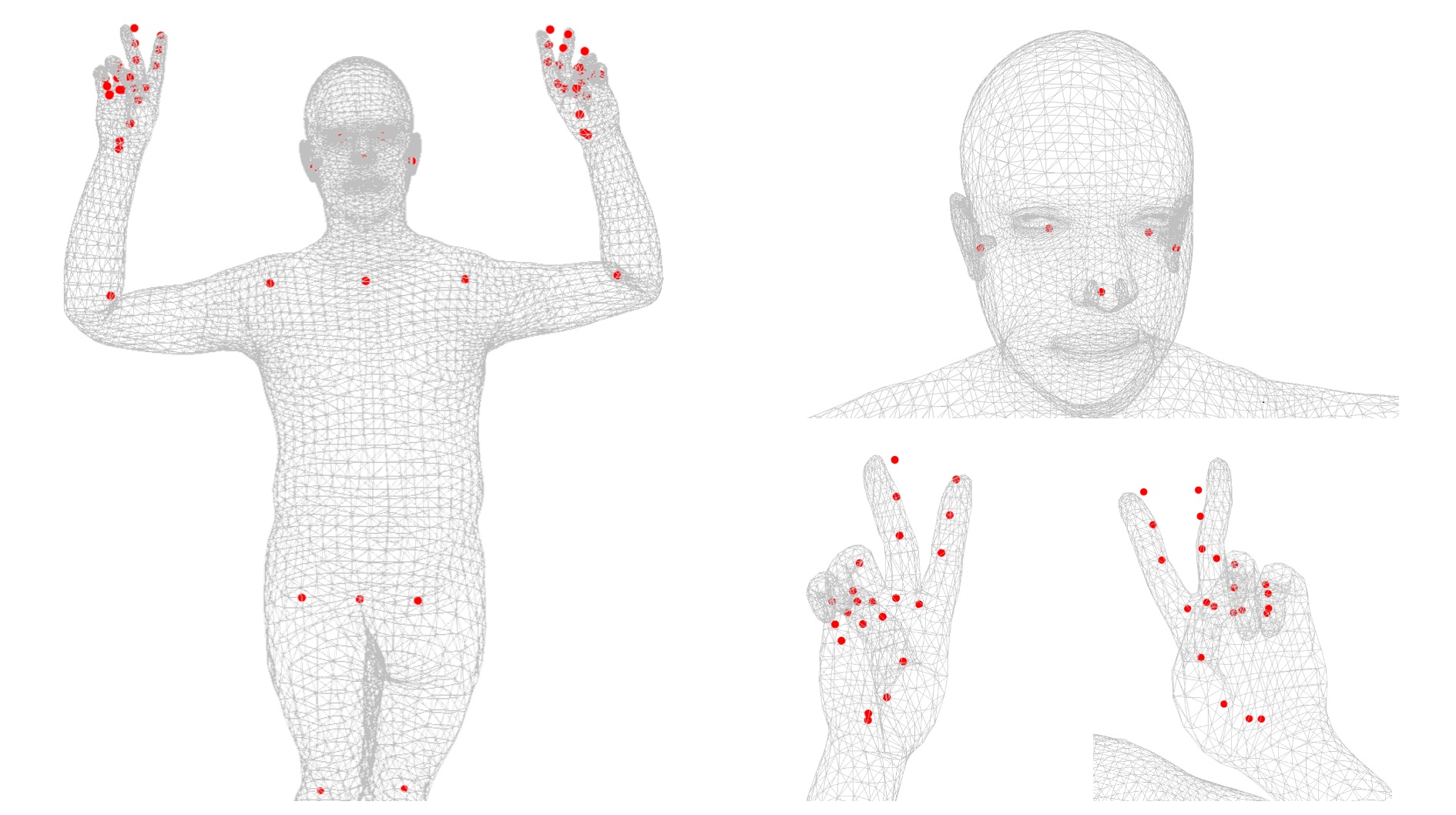}
  \caption{Comparison of our sparse-view method(8 view used) to ground truth from Total Capture Dataset\cite{joo2018total}. The mesh refers to our reconstructed parametric model (SMPL-X), and the red keypoints are the ground truth from the Total Capture dataset.}
  \label{fig:eval_total}
  \vspace{-3mm}
\end{figure}

Since our method is the first to enable lightweight total capture from sparse multi-view, we compared our method with SOTA single view method FrankMocap~\cite{Frankmocap} in Fig.~\ref{fig:single_qualitative} and ground truth from Total Capture~\cite{joo2018total} in Fig.~\ref{fig:eval_total}. 
What's more, MPJPE (Mean Per Joint Position Error) are provided for Total Capture dataset Tab.~\ref{tab:quantitative_cmu}.

\subsection{Evaluation: Hand Bootstrapping}
\label{sec:qualitative}

\begin{figure}
  \centering
  \includegraphics[width=\linewidth]{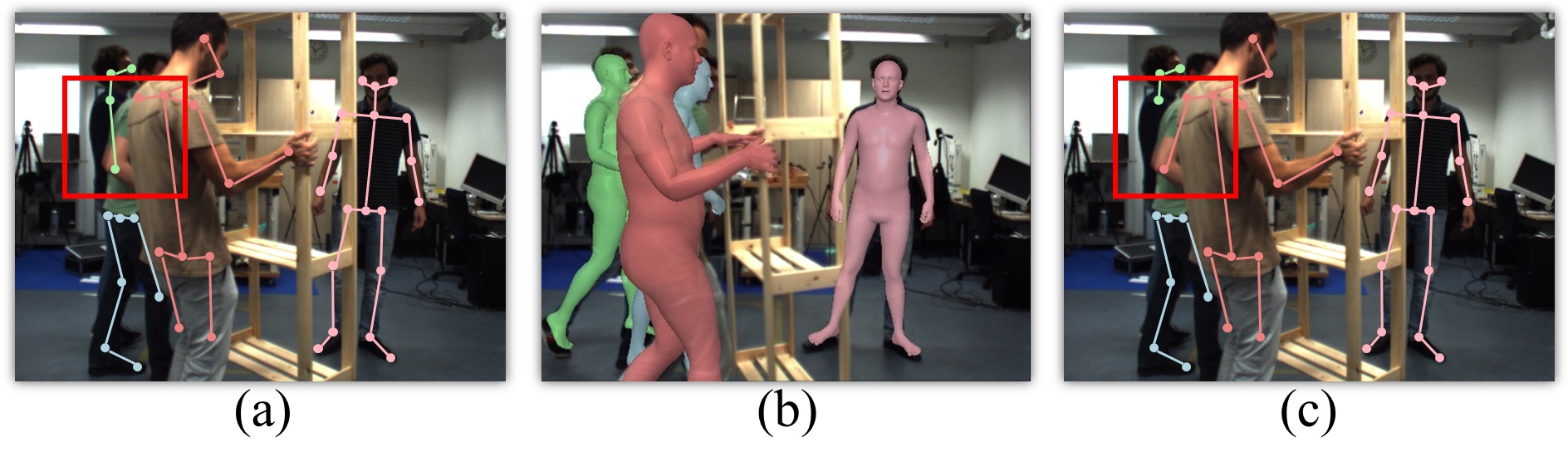}
  \caption{Qualitative evaluation of feedback mechanism. (a) shows the original association results of \cite{4DAssociation}. (b) are our reconstructed model of last frame. (c) shows that our feedback mechanism boosts body association performance with visibility prior. }
  \label{fig:qual_feedback}
  \vspace{-3mm}
\end{figure}

\begin{figure}
  \centering
  \includegraphics[width=0.95\linewidth]{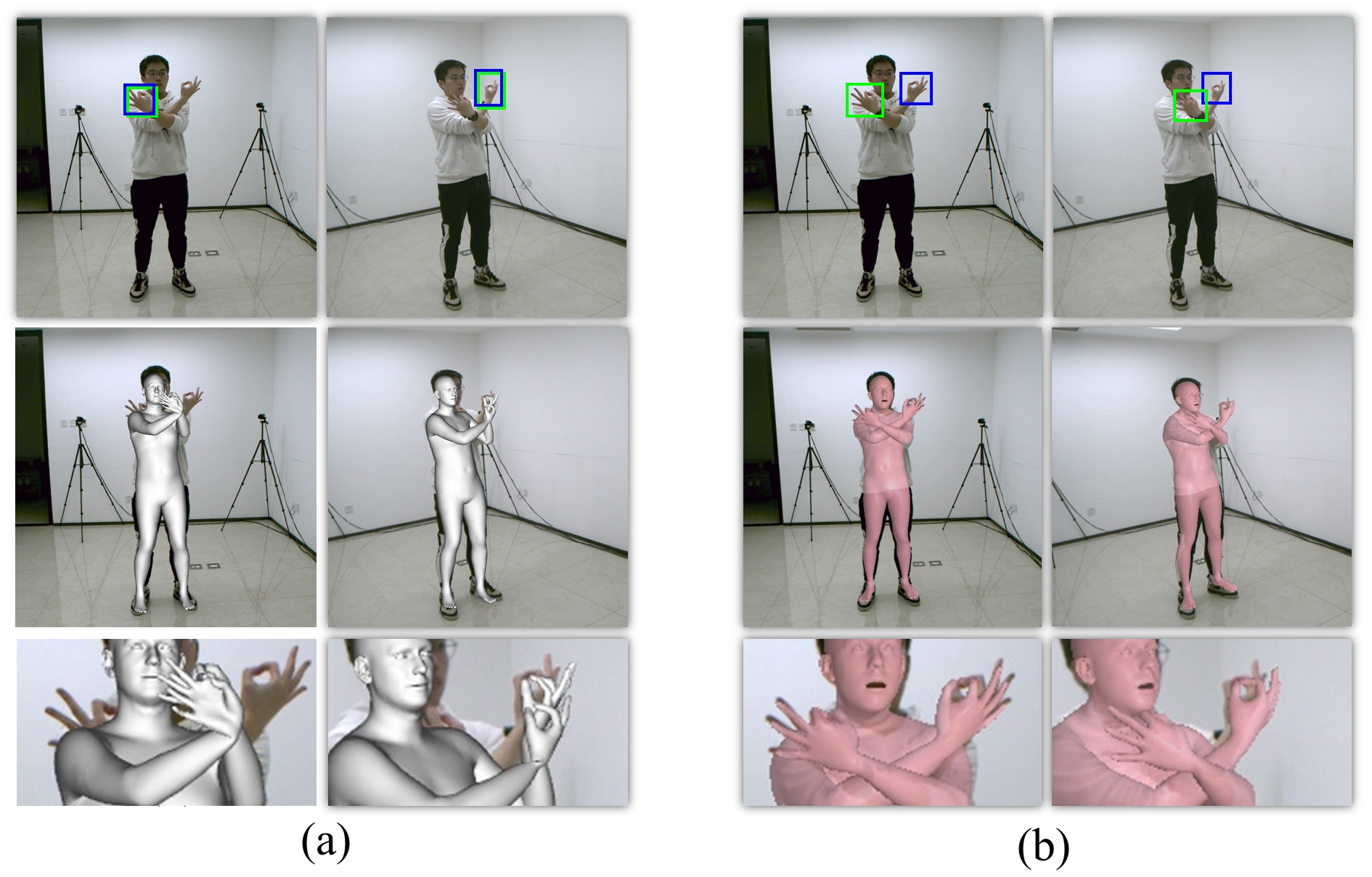}
  \caption{Qualitative evaluation of hand bootstrapping \& comparison against SOTA monocular method, FrankMocap \cite{Frankmocap}. (a) Results of Frankmocap \cite{Frankmocap}, only single ROI are extracted for each view, and left hand (blue rectangle) and right hand (green rectangle) have been distributed to the same ROI proposal. (b) Results of our method, all hands are extracted and associated correctly.}
  \label{fig:single_qualitative}
  \vspace{-3mm}
\end{figure}

We compare our hand bootstrapping method with SOTA monocular total capture method Frankmocap \cite{Frankmocap}. To ensure fairness as much as possible, we reduce our system to 2 close front view camera. 
Fig.~\ref{fig:single_qualitative} (a) shows the reconstruction failure of FrankMocap \cite{Frankmocap} caused by mixing up left and right hands to the same region proposal. Thanks to the proposed NMS method in hand association, our method can robustly reconstruct more accurate hands in Fig.~\ref{fig:single_qualitative} (b). 

\subsection{Evaluation: Two-stage Parametric Fitting}
\label{sec:quantitative}
We conduct ablation study of two-stage fitting metric on CMU dataset \cite{xiang2018monoculartotal} and demonstrate that our method makes different modality detectors benefit from each other. On the one hand, as shown in Fig.~\ref{fig:qual_fitting} (a)(d), we perform our two-stage fitting algorithm with only pose regression results, namely we leverage orthogonal projected joints from pose-regression network to take over heatmap-based 2D correspondences in stage two. Misalignment artifacts are shown in detail as pose-regression detector could not guarantee pixel-aligned accuracy. One the other hand, keypoints-detection-only results are shown in Fig.~\ref{fig:qual_fitting} (b)(e). Without pose regression network to initialize hand poses with reasonable gestures, the optimization is easy to fall into a local minimum.

\begin{table}[ht]
\centering
\begin{tabular}{lllll}
\hline
\quad Type &Body &Head &LHand &RHand  \\ \hline
MPJPE(mm) &33.4 &21.7 &22.6 &19.3  \\\hline
\end{tabular}
\vspace{1mm}
\caption{Quantitative evaluation on Total Capture dataset. We calculate the MPJPE of body\&head joints on a video segment (750 frames) which involves both large variation movements and meticulous hand gestures (noted that hand annotations are few for challenge gestures) with 5 cameras for a comprehensive evaluation.}
\label{tab:quantitative_cmu}
\end{table}


\subsection{Evaluation: Feedback Mechanism}
\label{sec:ablation}

We evaluate our feedback module in Shelf dataset~\cite{belagiannis20143d}. As shown in Fig.~\ref{fig:qual_feedback}(a), the left elbow the of salmon person is distributed to the background green person without feedback. 
We show enhanced association results in Fig.~\ref{fig:qual_feedback}(c) and prove that visibility informations provided by reconstructed human models help to eliminate that ambiguity.

\begin{table}[ht]
\centering
\begin{tabular}{lllll}
\hline
\quad Shelf &\quad A1 &\quad A2 &\quad A3 &\quad Avg \\ \hline
w/o feedback &\quad $99.0$ &\quad $96.2$ &\quad $97.6$ &\quad $97.6$\\
w/ feedback &\quad $99.5$ &\quad $97.0$ &\quad $97.8$ &\quad $98.1$ \\ \hline
\end{tabular}
\vspace{2mm}
\caption{Ablation study of feedback mechanism on Shelf dataset. Numbers are percentage of correct parts(PCP).}
\label{tab:ablation_feedback}
\end{table}

\vspace{-5mm}
\section{Discussion}
\label{sec:conclusion}
\noindent\textbf{Conclusion} In this paper, we propose, as far as we know, the first multi-person total motion capture framework with only a sparse multi-view setup. Based on the proposed hand and face bootstrapping, two-stage parametric fitting and feedback mechanism, our method can enable lighweight, fast, robust and accurate capture of the body pose, hand gesture and facial expression of each character even under the scenarios with severe occlusions and close interactions.

\noindent\textbf{Limitation and Future Work} We can mainly recover the facial expression by the jaw joint, and cannot reconstruct subtle facial expressions due to the low-resolution facial image input, which we leave for future research.

\noindent\textbf{Acknowledgements} This work is supported by the National Key Research and Development Program of China No.2018YFB2100500 and the NSFC No.62125107, No.61827805, No.62171255, and the Shuimu Tsinghua Scholarship.  

\clearpage
{\small
\

\bibliographystyle{ieee_fullname}
}

\end{document}